\documentclass[preprint]{JASA}

\usepackage{graphicx}
\usepackage{booktabs}
\usepackage{multirow}
\usepackage{expex}
\usepackage[margin=1in]{geometry}
\usepackage{array}
\usepackage{tipa}
\usepackage{rotating}
\usepackage{comment}

\usepackage{xeCJK}
\usepackage{lineno}

\begin{document}

\title[Direct cueing of word identity through word duration: evidence from Mandarin homophones]{Direct cueing of word identity through word duration: evidence from Mandarin homophones}



\author{Xiaomeng Zhu}
\affiliation{Department of Linguistics, Yale University, New Haven, USA}

\author{Megan Niu}
\affiliation{Department of Linguistics, Yale University, New Haven, USA}

\author{Jason Shaw}
\affiliation{Department of Linguistics, Yale University, New Haven, USA}

\author{Kevin Tang}
\email{kevin.tang@hhu.de}
\altaffiliation{Corresponding author. Also at: Department of Linguistics, University of Florida, Gainesville, Florida, 32611-5454, United States of America}
\affiliation{Department of English Language and Linguistics, Institute of English and American Studies, Faculty of Arts and Humanities, Heinrich Heine University Düsseldorf, Düsseldorf, 40225, Germany}




\begin{abstract}
\nolinenumbers
While there is increasing evidence that words are stored in the mental lexicon with some phonetic detail, there are open questions about exactly which phonetic details are stored and whether they are functionally relevant for spoken word recognition. Word-level phonetic cues, such as word duration, provide an appropriate test case. Across languages, including Mandarin Chinese, high frequency words are shorter than low frequency words. Using a homophone ordering task, we investigate whether differences in word duration impact word recognition. Mandarin Chinese was examined since it has abundant disyllabic homophones.
Our results indicate that participants were able to use word duration to recognize the order of homophone pairs, selected to vary in lexical frequency. Both with high noise levels and when no other cues were available, participants mapped the shorter duration auditory stimulus to the higher frequency word. The results provide evidence that word duration can directly cue the identity of words, suggesting that it may be a phonetic detail incorporated into the lexical representation. We discuss this direct cueing account alongside alternative accounts of how word duration may modulate word recognition. 

\end{abstract}


\maketitle
\nolinenumbers 



\section{\label{sec:intro} Introduction}

There is ample evidence that the temporal duration of linguistic units in the speech signal may function to cue the identity of those units. This is true for languages with phonological length, such as Japanese, where the duration of segments (vowels and consonants) alone can differentiate meaning: e.g., \textit{kite} `come (imp.)' vs.\ \textit{kiite} `listen (imp.)' vs.\ \textit{kitte} `postage stamp'. It is also true for languages without phonological length, such as English, where vowel duration is a robust cue to certain contrasts, including tense/lax pairs, such as [i] vs.\ [\textipa{I}], vowel height pairs, such as [\textipa{E}] vs.\ [\textipa{ae}], and fricative voicing, [s] vs.\ [z]. In these cases, temporal duration facilitates lexical recognition because it cues the identity of the phonological units that make up a word.

It is less clear whether temporal duration also cues higher-level linguistic units such as words. In speech production, word duration varies with several factors, including frequency, local predictability, and prosody. Whether these differences in word duration facilitate word recognition, however, remains an open question. Existing perceptual evidence is mixed and depends on the task and the available context. In a cloze task, manipulations of duration had little effect on word recognition compared with manipulations of pitch and spectral properties \citep{Bernhard_Tucker_2015,Podlubny_et_al_2018}. Other work suggests that duration matters by shaping how listeners interpret the surrounding signal. For example, duration may affect the processing of reduced speech by modulating rhythmic expectations, specifically altering the speech rate of a context can make words appear or disappear \citep{dilley2010altering}.  When context is absent, duration manipulations have been reported to degrade word recognition accuracy \citep{vitevitch2002naturalistic}. Evidence that listeners may use durational cues in word retrieval can also be found in speech misperception studies \citep{TangNevins_2014_SegLexTrends_NEL43,Tang2015,Felty2013}. These studies have found a positive correlation between the lexical frequency of the target word and that of the perceived word, suggesting that durational information may influence which lexical candidates listeners retrieve. Taken together, these findings establish that duration can influence word recognition, but they leave open whether it does so only indirectly (e.g., by giving listeners more time to use context or to process other cues) or whether it can directly cue word identity.

Homophones offer a way to address this question while controlling for segmental influences on word duration. Because homophones share the same segmental (and, in Mandarin, tonal) makeup, any systematic difference between them in the signal cannot be attributed to their segments. A growing body of corpus-based studies of spontaneous speech (on at least English, Mandarin, and German) suggests that such differences may exist, because words carry word-specific phonetic properties. In English, \citet{Seyfarth_2014} found that word duration is potentially lexicalized and is related to a word's informativity, that is, its average contextual predictability. \citet{Tang_Shaw_2021} extended this finding to the three major prosodic cues (duration, intensity, and pitch) in Mandarin Chinese. Further evidence of word-specific duration and pitch realizations in English and Mandarin show that these phonetic values are partly predictable from the words' meanings in utterance context \citep{ChuangICPHS2023,jin2026usingembeddingspredictspoken,chuang2025wordspecifictonalrealizationsmandarin,lu2025realizationtonesspontaneousspoken}. Word-specific phonetics is not limited to single words: there is also evidence for the lexicalization of intonation (f0 contours) in English and German collocations \citep{CalhounSchweitzer+2013+271+328,SCHWEITZER201565}. These findings suggest that homophones may differ phonetically as well. In English, higher-frequency (more predictable) words tend to be shorter, so homophones that differ in frequency are expected to differ in duration even when their segments are identical \citep{Gahl_2008,Lohmann2018}. In Mandarin Chinese, \citet{Tang_Shaw_LSA_Jan2021} found that the duration of homophones is related to their informativity, such that a word of a homophonic set tends to be shorter than its other homophonic member if the former has a lower informativity (on average less predictable). Beyond duration, \citet{tseng2026lettershidetrutheyes} showed that English homophones also differ in other phonetic details and that these differences are predictable from the words' meanings in utterance context.

Whether listeners use such differences to recognize words and to distinguish homophones is less well understood. \citet{Sanker_2019} found that listeners' ability to distinguish homophones depended on the production context: listeners performed above chance when discriminating homophone pairs taken from sentences but not when the words were produced in isolation. This suggests that listeners can exploit acoustic details of homophones. Furthermore, English-speaking listeners were able to discriminate homophones when they differed in vowel duration, but not when they differed in mean F0 \citep{Sanker_LSA_2020}. \citet{Sanker_Homophone_JoP_2022} found that listeners can discriminate homophones at above-chance accuracy after prior exposure to the same test words, an effect that persisted across speakers. These studies highlight that listeners do not rely on just any acoustic detail. 

Evidence that duration is informative for homophones can be also found in computational work. \citet{schuppler22_interspeech} found that function words with homophonous realizations (\textit{ah, ach, auch, eine}, and \textit{er} can also be reduced to [a]) could be disambiguated better by a machine classifier when durational information was available. Similarly, \citet{SchupplerHomophone2018} studied the German forms \textit{der, die, das} and their inflections \textit{des, dem, den}, each of which can function as a determiner, relative pronoun, or demonstrative pronoun, and examined 193 acoustic features for automatic disambiguation. They found that durational features were particularly robust as nearly all of them survived feature selection. However, just because a cue is robust for machine learning does not necessarily mean that humans will be sensitive to it. A similar line of computational work on vowel distinctions in English shows that machine classification shows a higher sensitivity to duration cues than native speakers, unless the machine learning is explicitly directed towards spectral cues \citep{jurov2023speech}.
These findings leave a gap. That duration is useful to a machine classifier does not show that human listeners use it, and it is not clear whether the perceptual system is sensitive to word-level durational differences of the size found in production. Mandarin provides a good test case, since it has abundant homophones, especially among disyllabic words.

In this paper we ask whether listeners can use word duration alone to identify homophones that differ in lexical frequency. Our hypothesis is that, if listeners have learned the association between frequency and duration, they should map the shorter token of a homophone pair to the higher-frequency member and the longer token to the lower-frequency member. We situate this hypothesis against three ways in which duration could facilitate word recognition: (i) by giving more time for sentential context to support prediction, i.e., ``context integration"; (ii) by giving more time to process other cues to word identity, i.e., ``spectral integration" and (iii) by directly cuing word identity, i.e., ``direct cueing". 

Our study is designed primarily to distinguish account (iii), ``direct cueing", from accounts (i) and (ii). We present our stimuli in a neutral context. Because of this, the ``context integration" account predicts that listeners will not reliably associate the shorter/longer duration audio stimulus with the higher/lower frequency homophone. We also control our stimulus items to differ only in duration, holding other acoustic cues constant across homophone pairs. The ``spectral integration" account predicts that, in the general case, associations between word duration and frequency will facilitate word recognition only when non-durational cues are also present.  That is, if duration primarily facilitates word recognition by giving more time to process other cues, then masking cues will eliminate otherwise facilitatory effects of longer duration. In the case of Mandarin homophones, there may be other non-durational cues to word identity, such as pitch \citep{chuang2025wordspecifictonalrealizationsmandarin}. 
On the spectral integration account, duration could help to make subtle acoustic cues more salient to the listener. Our design rules out this possibility in this particular experiment. To focus on duration alone as a cue to word identity, we control all other cues differentiating homophone pairs. Under these conditions, the ``spectral integration" account predicts the same outcome as the ``context integration" account, listeners should not be able to associate word duration differences with homophone frequency. Direct cueing, in contrast,  predicts an effect of word duration on word recognition that persists even when no other cues are available. Notably, our design is narrowly focused on testing the direct cueing account while controlling for factors that would be relevant for ``context integration" and ``spectral integration" accounts, which may still be relevant in other contexts.

\begin{table}[ht]
\centering
\begin{tabular}{lccc}
\toprule

Account & (i) Context integration & (ii) Spectral integration & (iii) Direct cueing \\
\midrule
Prediction   & At chance  & At chance & Above chance \\
\bottomrule
\end{tabular}
\caption{Predictions of the three accounts for our experiments testing homophone identification based on word duration.}
\label{tab:account_predictions}
\end{table}

We report two experiments. In Experiment 1 (Section \ref{sec:exp1}), we manipulated the duration of Mandarin homophone pairs embedded in a carrier phrase and masked with noise, so that duration was the only cue distinguishing the members of each pair. In Experiment 2 (Section \ref{sec:exp2}), we removed the speech signal altogether, leaving only noise of the same duration, to test whether the effect persists in the absence of an intelligible signal. Of the three accounts discussed above, only direct cueing predicts above-chance behavior in these experiments.

\section{\label{sec:exp1} Experiment 1: Some signal for homophone pairs}

\subsection{\label{subsec:methods} Methods}

\subsubsection{Design}

In this experiment, we test whether participants are sensitive to duration as a cue for identifying homophones that differ in lexical frequency. More specifically, given a pair of homophones that differ in frequency, we ask whether they were able to associate the high-frequency word with shorter duration and its low-frequency counterpart with longer duration.

\begin{figure}
    \centering
    \includegraphics[width=0.8\linewidth]{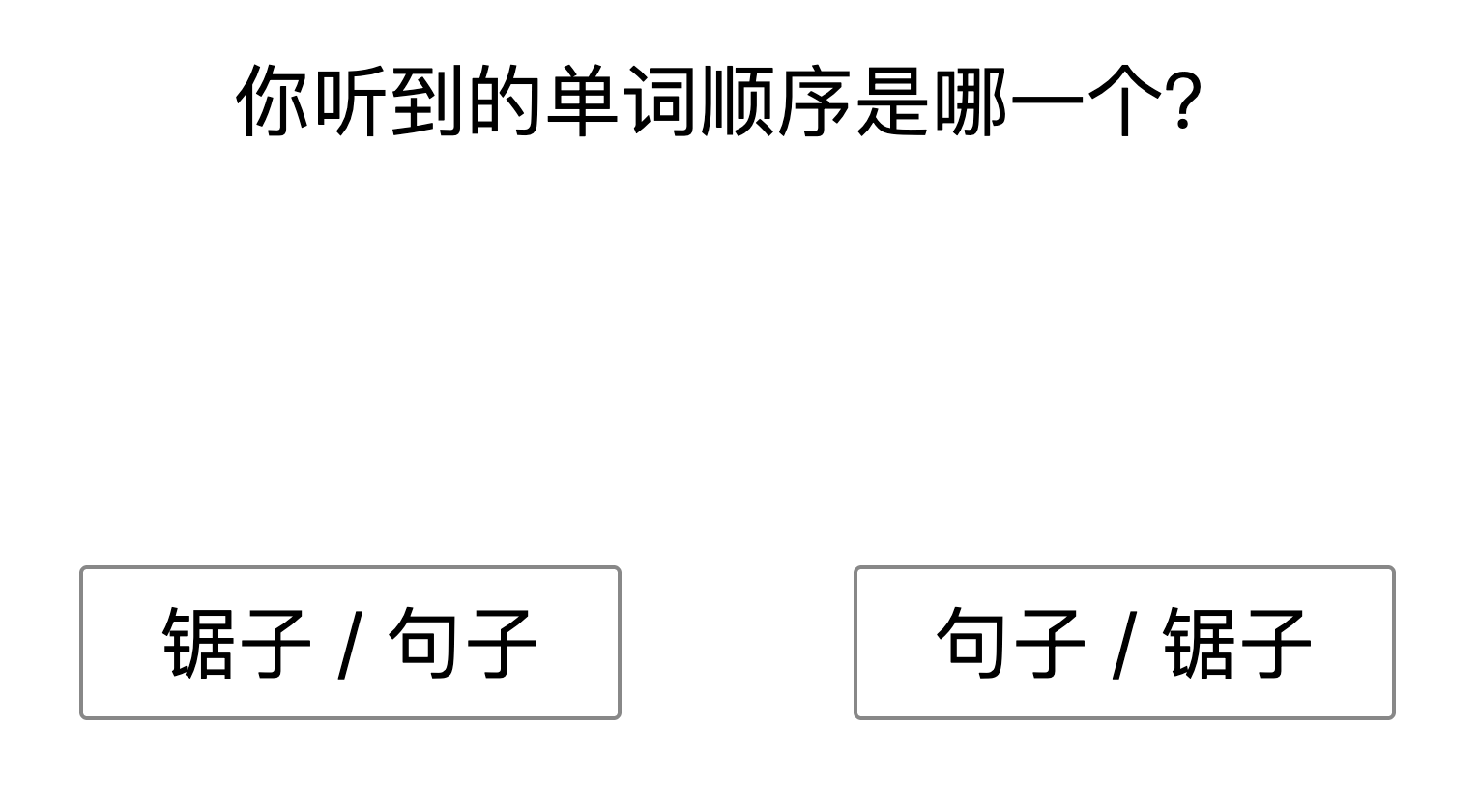}
    \caption{Experimental interface in Gorilla showing an example test item (homophone pairs). The instruction at the top says ``Which order of words did you hear?''. The left button has a homophone pair of low-high order (frequency), and the right button shows the reverse order.}
    \label{fig:interface}
\end{figure}

To achieve this goal, we present participants with audio stimuli that contain homophone pairs that were masked with noise in carrier phrases. The only difference between the auditory signals for the target words in the pair is duration. In some conditions, the duration of the first target noise is shorter than the second (henceforth \textit{short-long}). In others, the first is longer than the second (henceforth \textit{long-short}). Participants were then asked to pick the order of the homophone they heard. They were presented with two response buttons that differ in the order of the homophone pair. One response button shows the high frequency word first, and then the low frequency one (henceforth \textit{high-low}), while the other button shows the reverse order (i.e., \textit{low-high}). For example, one button is 句子 (`sentence') / 锯子 (`saw'), where the first word is more frequent than the second one. In this case, the other button would be 锯子 (`saw') / 句子 (`sentence'), where the first word is less frequent than the second. An example of our experimental interface is shown in Figure \ref{fig:interface}.

If word duration alone can serve as a cue for homophone perception, then participants will be more likely to pick the word order whose frequency corresponds to the duration cue. Specifically, if they hear \textit{long-short}, they should be more likely to pick the button with the frequency \textit{low-high}, since low-frequency words tend to be longer than high-frequency ones. On the other hand, if the audio is \textit{short-long}, they should be more likely to pick the button with the frequency \textit{high-low} than \textit{low-high}.

\subsubsection{Materials}

The homophone identification task consisted of 48 trials in total, with 16 target trials and 32 filler trials.
All words were produced in the carrier phrase in (1). 
For target trials, the words in the blanks are homophones. For filler trials, the words constitute phonetically-similar but non-homophonous pairs. For example, one word is 道路 (dào lù; `road') and the other is 大路 (dà lù; `avenue').
The words in the pair also differ in frequency.

\ex
\begingl
\gla diyige ci shi \_\_\_\_\_ dierge ci shi \_\_\_\_\_ xing ma? //
\glb first word \textsc{be} \_\_\_\_\_  second word \textsc{be} \_\_\_\_\_ correct \textsc{sfp}//
\glft The first word is \_\_\_\_\_, the second word is \_\_\_\_\_, is that correct? //
\endgl
\xe

\paragraph{\label{subsubsec:stimuliselection} Stimuli selection}
Target and filler stimuli were selected from the Chinese Lexical Database (CLD) \citep{Sun_et_al_2018}.

The target stimuli were identified using six criteria. First, each item belonged to a homophone set with exactly two members (i.e., a pair of homophones rather than a set of three or more homophones). Second, both items in the pair were disyllabic. This is because disyllabic words were the most frequent word length in Mandarin \citep{Neergaard2022} as well as the word type that includes the most homophones. Third, both items (i.e., high and low) in the pair had the same part of speech. This controls for possible confounding effects for different parts of speech, such as noun-verb duration effects \citep{LOHMANN_2018,LOHMANN2020_JoP,Lohmann_2020_Linguistics}. Fourth, the dominant part of speech accounted for at least 90\% of occurrences for each member of the pair. Fifth, the prevalence of both items was greater than 80\%\footnote{Over 80\% of participants correctly identified these items in a  lexical decision task.} to ensure familiarity to speakers, especially the low-frequency words \citep{Keuleersetal2015}. Sixth, the pair has a cosine similarity less than 0.5 to ensure that the environments in which the words occur are sufficiently different, so that their meanings are sufficiently different and not near-synonyms \citep{Mikolov_et_al_2013}.

After applying the criteria listed above, we obtained 143 eligible homophone pairs. From these candidate pairs, we selected 16 target pairs using the Zipf scale of word frequency \citep{Brysbaert2018} (with 1 being least  frequent, and 7 being most frequent). The resulting set of target stimuli contained high-frequency words with Zipf values ranging from 3.8 to 4.8, as well as low-frequency words with Zipf values ranging from 2.0 to 3.0. The mean difference in Zipf frequency between the high- and low-frequency words was 1.58.

In terms of filler items, we applied criteria 2-6 from the target stimuli selection process. Additionally, we restricted to words that began with voiceless stops (either aspirated or unaspirated) that did not belong to any homophone set. For each target homophone pair, we selected two filler items that were minimal pairs with similar lexical variables.

\paragraph{\label{subsubsec:stimulirecordingandprocessing} Stimuli recording and preprocessing}

After selecting the target and filler stimulus items, the stimuli were recorded by a male native speaker of Mandarin Chinese in his 20's at the time of recording. Six repetitions of each pair were recorded.
We excluded the repetitions whose low-frequency member had a duration greater than 2.5 standard deviations from the mean duration across items. 
We then selected the token whose low-frequency member had the longest duration. 
Correspondingly, the high-frequency token of the pair was manipulated such that its duration was 25\% shorter than that of its low-frequency counterpart. This was to approximate the expected effect of lexical frequency on word duration. 
Duration manipulation was performed using a custom Python script, which interfaced with Praat's \citep{praat} Manipulation Object through the \texttt{parselmouth} library \citep{JADOUL20181}.
Importantly, the duration of the high-frequency token was manipulated, while all other properties of the recorded signal was preserved.

After duration manipulation, we imposed speech-shaped noise on the test items in the sentence. 
The duration of the noise matched the manipulated duration of the words. In other words, since high frequency words were shorter than that for low frequency words in the audio, the noise was also correspondingly shorter. There were two different noise levels: low (0 dB) and medium (-8 dB).
Under the low noise condition, the words were expected to remain readily identifiable. For the medium-noise condition, the identification was expected to be more difficult.

In the current experiment, the homophone pairs in the target trials were masked with the low noise level (0 dB). Since the homophone pairs were segmentally identical and differed only in their manipulated durations, this design left duration as the only available cue for distinguishing them.


\subsubsection{\label{subsec:Procedure} Procedure}
200 native Mandarin speakers participated in the experiment. Half were recruited on Prolific, and the remaining half were recruited on the Chinese social media platform Xiaohongshu.
All participants were directed to the experiment-building platform Gorilla to complete the experiment. There were three separate tasks in the experiment: (i) homophone identification, (ii) LexTale, and (iii) a demographic questionnaire.

First, homophone identification is the main task. On each trial, participants were first briefly (for 500 ms) presented with the two response options before being presented with the auditory stimulus. This was intended to encourage the participants to attend to both response alternatives before hearing the stimulus.
We adopted this priming procedure following \citet{Sanker_2019}, who found that simultaneous presentation of the response options and auditory stimulus increased a positional bias that facilitates the choice of the first orthographic form that is compatible with the speech signal. An example test trial was shown in Figure \ref{fig:interface}.

After the preview, the auditory stimulus was played while both response options remained visible on the screen. The order of the response options were randomized across trials (i.e., the left button could have \textit{high-low} first or \textit{low-high} first). Participants could not click on the buttons to advance to the next trial until the audio had finished playing and the question appeared on the screen. The audio was only played once with no replay option. The inter-trial interval was set to 500 ms. 

\ex
\begingl
\gla ni tingdao de danci shunxu shi na yige? //
\glb you hear \textsc{de} word order \textsc{be} which one? //
\glft Which order of words did you hear? //
\endgl
\xe

Participants received two practice trials, one from the low noise level and one from the medium noise level, and the words in the practice trials were minimal pairs of each other.
As described in the previous section, there were 48 experimental trials (16 tests and 32 fillers), and the homophones in the target trials were masked with the \textbf{low} noise level.
The words in the filler trials were randomly assigned to either the low or medium noise level. Given the difficulty of the test trials, we wanted the easier filler trials to serve as encouragement through the task. 
Each of the 32 filler trials had eight possible conditions—for each condition below, there was both a low and medium noise level option. The condition chosen for each trial was randomized across participants.

In the second task, participants completed the LexTALE task as a measure of their lexical knowledge. Participants were asked whether or not the character on the screen was a real simplified Mandarin character using a web-compatible version of LexTALE\_CH \citep{Chan_Chang_2018} by \citet{tang_lextale_ch_upscaled_2024}. 
The LexTALE task consisted of 60 lexical items (i.e., real Chinese characters) and 30 nonce items (i.e., non-words). Participants were visually presented with 90 items total.

In the last part of the experiment, participants completed a demographics questionnaire that collected information about age, gender, and language background.

\subsection{\label{sec:Results} Analysis}

The dependent variable was participants' choice of homophone order, coded as either \textit{high-low} (i.e., choosing the order where the high-frequency item is followed by the low-frequency item) or \textit{low-high} (the low-frequency item followed by the high-frequency item). We fitted the binary responses using logistic mixed-effects regression implemented with the \texttt{glmer} function in the \texttt{lme4} package \citep{lme4} in R (\citealp{R}, version 4.3.1). In the regression model, \textit{low-high} was treated as the reference level.

The main fixed-effect predictor of interest was the duration pattern of the auditory stimulus (\textit{long-short} vs.\ \textit{short-long}). The prediction is that if the auditory stimulus was \textit{short-long}, participants should be more likely to pick the response of \textit{high-low}. On the other hand, if the auditory stimulus was \textit{long-short}, participants should be more likely to pick the response of \textit{low-high}.

In the analysis, the variable auditory stimulus type was sum-coded, with \textit{long-short} coded as $-0.5$ and \textit{short-long} as $0.5$. Trial number (i.e., the order of the trial in the presentation sequence) was standardized and included as a fixed-effect predictor to account for changes in responses over the course of the experiment. The model also included random intercepts for participant and target item. Only target trials from participants who passed the lexical proficiency screening were included in the analysis.

More explicitly, the structure of the logistic mixed-effects model is as follows: \texttt{Frequency.Chosen $\sim$ Auditory.Stimuli + Trial\_z + (1 | Participant.ID) + (1 | Trial.ID)}.

\subsection{Results}


We first analyzed the performance of participants on the supplementary LexTALE task in order to exclude participants whose Mandarin proficiency was insufficient for the purposes of our experiment. 164 out of 200 participants passed the LexTale task using a cutoff of 0.73 in raw accuracy and 28 in corrected accuracy. 

For the remaining participants, we analyzed their accuracy in the filler items. First, the filler accuracy in the low noise condition is higher than the high noise condition, which suggests that the noise manipulation affected participants' perception of non-homophone filler pairs. This difference in noise conditions is confirmed by a logistic mixed-effect model predicting correctness on the filler trial with noise level and trial order as fixed-effect predictors as well as participant and item as random intercepts ($\beta = -2.01$, $SE = 0.11$, $z = -17.53$, $p < .001$).
Importantly, participants recruited from both Prolific and Xiaohongshu achieved comparable filler accuracy across noise conditions.
In the low-noise condition, accuracy was high for both Prolific (96.7\%) and Xiaohongshu (95.3\%) participants. In the high-noise condition, accuracy decreased for both groups as expected, reaching 82.7\% for Prolific participants and 75.2\% for Xiaohongshu participants.
This suggests that participants recruited through both sources performed similarly on the filler trials, providing no evidence of a systematic difference in participant quality between the two samples.

We next examined whether the duration can serve as a cue for homophone perception. We fitted the logistic regression model described in the previous section, with auditory stimulus type and standardized trial number as fixed-effect predictors and random intercepts for item and participant.\footnote{The model with the participant-level random intercepts resulted in a singular fit. However, removing the participant-level random intercept and only keeping the item-level intercept produced virtually identical fixed-effect estimates, standard errors, and significance levels. We therefore report the results from the original model.}
Results are presented in Table~\ref{tab:exp1_model}.

\begin{table}[ht]
\centering
\begin{tabular}{lrrrrr}
\toprule
Parameter & Log-Odds & SE & 95\% CI & $z$ & $p$ \\
\midrule
Intercept
    & 0.11 & 0.04 & $[0.03,\ 0.19]$ & 2.81 & 0.005 \\

Auditory Stimuli [Short--Long]
    & 0.24 & 0.08 & $[0.08,\ 0.39]$ & 3.01 & 0.003 \\

Trial $z$
    & 0.01 & 0.04 & $[-0.07,\ 0.09]$ & 0.26 & 0.793 \\
\bottomrule
\end{tabular}
\caption{Results of the logistic mixed-effects regression model predicting homophone-order choice as a function of auditory stimulus type and standardized trial order in Experiment 1.}
\label{tab:exp1_model}
\end{table}

Crucially, auditory stimulus type is a significant main effect predictor of word order selection ($\beta = 0.24$, $SE = 0.08$, $z = 3.01$, $p = .003$). This suggests that participants were more likely to select the \textit{high-low} ordering when the duration of the first homophone item is shorter than the second. This pattern is consistent with the predicted mapping between frequency and duration. Additionally, trial order did not significantly predict response choice ($p = .793$), which suggests that the response patterns did not change throughout the course of the experiment.

\section{\label{sec:exp2} Experiment 2: No signal for homophone pairs}
In Experiment 1, we found that participants' choice of homophone order was influenced by the duration pattern of the auditory stimulus: participants were more likely to choose the \textit{high-low} ordering when the auditory stimuli were \textit{short-long}. This suggests that listeners were able to use duration as a cue for identifying homophones that differ in lexical frequency. This is especially striking given that both items in the pair are segmentally identical and only differ in (manipulated) duration.
In Experiment 2, we took this one step further and tested whether the effect would persist in the absence of an intelligible speech signal.
Specifically, the items in the homophone pair for target trials were presented under the highest noise condition. In this condition, the speech signal was unintelligible, and listeners could only perceive the noise with an intensity level of $65$dB with no signal present. The difference in duration between the high-frequency item and the low-frequency item in the pair remains. This allowed us to test in a more strict condition whether duration alone could serve as a cue for identifying homophones that differ in frequency.

\subsection{Methods}

Experiment 2 is the same as Experiment 1 except for the following two modifications. First, as described above, the speech-shaped noise imposed on the target items was of the highest noise condition (noise only with an intensity value of $65$dB). Second, we used the platform Prolific to recruit an additional 200 participants. Participants were recruited in two rounds (50 in the first, 150 in the second).

\subsection{Results}

Out of the 200 participants who participated in Experiment 2, 174 participants passed the LexTale task. In terms of fillers, they achieved an accuracy of 0.78 for high noise conditions and an accuracy of 0.95 for low noise conditions. This is comparable with the filler accuracies reported in Experiment 1.

We used the same mixed-effect model structure as Experiment 1. Regression results were shown in Table \ref{tab:exp2_model}. Auditory stimulus type remains as a significant predictor of the choice of homophone order ($\beta = 0.20$, $SE = 0.08$, $z = 2.60$, $p = 0.009$). As in Experiment 1, participants were more likely to select the \textit{high-low} ordering when the auditory stimulus was \textit{short-long}. Trial order also did not surface as a significant main effect ($p = .805$).

\begin{table}[ht]
\centering
\begin{tabular}{lrrrrr}
\toprule
Parameter & Log-Odds & SE & 95\% CI & $z$ & $p$ \\
\midrule
Intercept
    & 0.01 & 0.05 & $[-0.08,\ 0.10]$ & 0.25 & 0.804 \\

Auditory Stimuli [Short--Long]
    & 0.20 & 0.08 & $[0.05,\ 0.35]$ & 2.60 & 0.009 \\

Trial $z$
    & 0.00946 & 0.04 & $[-0.07,\ 0.08]$ & 0.25 & 0.805 \\
\bottomrule
\end{tabular}
\caption{Results of the logistic mixed-effects regression model predicting homophone-order choice as a function of auditory stimulus type and standardized trial order in Experiment 2.}
\label{tab:exp2_model}
\end{table}

To test whether noise level on the target items modulated listeners' use of duration as a cue to homophone identity, we fit a logistic mixed-effects model that includes noise level as a fixed-effect predictor in addition to the existing model structure. Crucially, noise level was allowed to interact with auditory stimulus type in order to test whether the effect of the \textit{short-long} versus \textit{long-short} duration pattern differed across noise conditions. The structure of the interactive model is as follows: \texttt{Frequency.Chosen $\sim$ Auditory.Stimuli * ExpVersion + Trial\_z + (1 | Participant.ID) + (1 | Trial.ID)}.

The regression results are shown in Table \ref{tab:exp2_model_combined}. The interaction between auditory stimulus type and noise level was not significant ($p = .700$), indicating that the effect of duration patterns on homophone-order choice did not reliably differ across noise conditions. Together with the significant effect of auditory stimulus type, this pattern suggests that the duration effect was detectable even in the absence of any will signal.

\begin{table}[ht]
\centering
\begin{tabular}{lrrrrr}
\toprule
Parameter & Log-Odds & SE & 95\% CI & $z$ & $p$ \\
\midrule
Intercept
    & 0.06 & 0.03 & $[0.00,\ 0.12]$ & 2.05 & 0.040 \\

Auditory Stimuli [Short--Long]
    & 0.22 & 0.05 & $[0.11,\ 0.32]$ & 3.94 & $< .001$ \\

ExpVersionBinary [easyNoise]
    & 0.10 & 0.06 & $[-0.01,\ 0.21]$ & 1.82 & 0.069 \\

Trial $z$
    & 0.010 & 0.03 & $[-0.04,\ 0.06]$ & 0.36 & 0.715 \\

Auditory Stimuli [Short--Long] $\times$
ExpVersionBinary [easyNoise]
    & 0.04 & 0.11 & $[-0.17,\ 0.26]$ & 0.39 & 0.700 \\
\bottomrule
\end{tabular}
\caption{Results of the logistic mixed-effects regression model predicting homophone-order choice from auditory stimulus type, experiment, their interaction, and standardized trial order. Categorical predictors were coded using $-0.5$ and $+0.5$ contrasts.}
\label{tab:exp2_model_combined}
\end{table}

\section{\label{sec:Discussion} Discussion}
The effect of word predictability on word duration is robust across languages, such as English \citep{Belletal2003,aylett2004smooth}, Dutch \citep{pluymaekers_articulatory_2005}, Japanese \citep{hashimoto_probabilistic_2021}, Mandarin Chinese \citep{Tang_Bennett_Kaq_Predictability_JASA_2018} and Kaqchikel \citep{Tang_Bennett_Kaq_Predictability_JASA_2018}. Words that are more predictable tend to be shorter in duration. Less is known about the relevance of such duration differences for speech perception. 

Generally speaking, increasing the duration of an interval could facilitate word recognition for a couple of different reasons, which we described in the introduction and summarize in TABLE \ref{tab:account_predictions} along with predictions.
Word duration could, as past work has suggested, facilitate word recognition by allowing more time for sentential context to facilitate prediction \citep{Podlubny_et_al_2018}. We refer to this account as ``context integration". A second way is by providing more time to process other (non-durational) cues to word identity, which we refer to as ``spectral integration". Finally, a third way that duration could influence perception is by providing a direct cue to word identity, which we refer to as ``direct cueing". Our study was designed primarily to test this last possibility.

Results indicate that word duration facilitates word recognition and, moreover, that differences in word duration are associated with word frequency. Listeners were able to use word duration to distinguish between homophone pairs of differing frequency, mapping the higher frequency member of the pair to the stimulus of shorter duration.     

In our experiments, listeners had no context to facilitate word recognition. They were forced to rely on duration as a cue, either in tandem with degraded spectral cues (Experiment 1) or with no spectral cues at all (Experiment 2). In both experiments, participants were able to use duration as a cue, identifying the word order with greater than chance accuracy. This indicates that there is more to word duration effects in perception than just allowing more time to process other cues (spectral integration) or for sentential context to facilitate prediction (context integration). The results support direct cueing. 

The other two accounts we considered, context integration and spectral integration, predict at-chance behavior in both of our experiments. This is because, in our design, we deprived participants of both contextual cues and other (non-durational) phonetic cues to word identity. Since duration was the only available cue to word identity, the spectral integration account predicts that more time to process other cues would be fruitless, leading to at-chance behavior. This should be the same whether there are some uninformative spectral cues (Experiment 1) or none at all (Experiment 2). The same is true for the context integration account. The neutral context of target word presentation meant that additional time to integrate the sentential context would not lead to increased performance. Thus, for our experiments, both alternative accounts predict at-chance performance. 

Since we found a similar level of above chance performance in both experiments, our design provides positive evidence that word duration can directly cue lexical items. Duration has a role in lexical access above and beyond providing more time for processing.  
That said, our results do not rule out these other possibilities. In naturalistic contexts, increased duration may, in addition to directly cuing lexical items, provide additional time to process other cues, both contextual and spectral, to word identity.  

The totality of our results are most consistent with the direct cueing account-- word duration directly cues word identity. Participants were able to distinguish homophone pairs, even when no other cues were present. 

Our results have implications for how lexical items are represented. Word duration is a type of phonetic detail that is not represented in strings of segments, a form of lexical representation in classic models of spoken word recognition, e.g., Cohort model \citep{MarslenWilson_Welsh_1978}, ShortList \citep{norris1994shortlist,Norris_McQueen_2008}, Trace \citep{McClelland_Elman_1986}. There is a consensus that the facts of speech perception necessitate a hybrid model of speech representation, combining both abstract categories and phonetic detail \citep{cutler2010abstraction, goldinger2007complementary,mcmurray2022myth, hawkins2003roles, baese2022nature}.
The precise mechanism for encoding phonetic detail, e.g., exemplars \citep{Pierrehumbert_2002}, memory buffers \citep{muegge2026decoupling}, dynamic neural fields \citep{kirov2007dynamic,shawtang_DFT_AMP2023}, neural networks \citep{beguvs2021ciwgan, beguvs2023encoding, magnuson2020earshot, scharenborg2010modeling}, and the precise dimensions of phonetic detail that are encoded remain frontier issues in speech perception. The current work contributes to the latter issue. Evidence that the phonetic duration of words can directly cue their identity in speech suggests that this aspect of phonetic detail is encoded in the lexicon.

\section{\label{sec:Conclusion} Conclusion}

This study asked whether listeners can use word duration alone to identify homophones that differ in lexical frequency in Mandarin Chinese, a language known for having a high number of homophones. In two experiments, Mandarin listeners heard homophone pairs in a neutral carrier phrase, with duration as the only difference between the two words. When the first word was shorter than the second, listeners were more likely to choose the response order in which the higher-frequency homophone came first. The effect held when the words were partially masked by noise (Experiment 1) and when no speech signal was present at all, only noise of matching duration (Experiment 2). Moreover, the magnitude of the effect did not differ reliably between the two noise conditions.

Because the design removed sentential context and held all non-durational cues constant, the results are difficult to explain through context integration or spectral integration alone. They are most consistent with a direct-cueing account, in which word duration is itself associated with word identity. This implies that listeners’ lexical representations encode word-level temporal detail, and that this detail is available for lexical access. The finding supports hybrid models of speech perception that combine abstract phonological categories with fine phonetic detail, and it identifies word duration as one dimension of that detail.

More broadly, these findings suggest that the temporal properties of words are part of what listeners know about them. If duration can cue word identity when no other information is available, then models of spoken word recognition need a way to represent word-level timing alongside segmental and tonal form, and the frequency-duration relationship documented in production appears to be something listeners can draw on in perception. This extends the case for phonetic detail in lexical representation beyond segment-level cues to the duration of whole words, and it raises the possibility that other word-specific properties documented in corpus work \citep{Tang_Shaw_2021}, such as pitch and intensity, are similarly available to listeners. The effects here were modest, and the stimuli came from a single speaker, 16 item pairs, and one fixed duration ratio, so the generality of the mechanism remains to be established. Future work should test graded duration differences, multiple speakers, and other languages, and should ask how direct cueing by duration combines with sentential context and spectral cues in natural speech, where, as the three accounts considered here are not mutually exclusive, all of these sources of information are likely to contribute to word recognition.
\begin{acknowledgments}

This work was partly conducted as part of the Bachelor of Science thesis of Megan Niu. We thank Zhenrui Zhang for assistance with data collection in China for Experiment 1. All errors remain ours.

\end{acknowledgments}

\section*{Author declarations}

\subsection*{Conflict of Interest}
There are no conflicts to disclose.

\subsection*{Ethics Approval}

The data collection was in accordance with Yale University's IRB No. 2000036833.

\section*{Data Availability}

The datasets generated during the current study are available in this published article and its supplementary information files; and in the \href{https://zenodo.org/records/23525789?preview=1&token=eyJhbGciOiJIUzUxMiJ9.eyJpZCI6Ijg2M2Q2ZjdiLTJlMWYtNDRhZC05MmJmLTVkYzdkNmRjZGVmMyIsImRhdGEiOnt9LCJyYW5kb20iOiI4OWQwNDVhYmI3YTE0M2FmZmUwOWM1YzRiNzgwYWQ5NiJ9.fLFmIR2T3uxw2pILrDrC3dxmYe6pHkA9ek_FNy5uGUD18IJpOzsYlxzOxYZNg5CpTtqnM3JvPakxd8z9q72AZw}{Zenodo repository}  (the repository is accessible via a hyperlink; it will be made public upon publication).

\section*{Authors' contributions}

Corresponding author: KT. We follow the CRediT taxonomy (\url{https://credit.niso.org/}). Conceptualization: XZ, MN, JS, KT; Methodology: XZ, MN, JS, KT; Investigation: XZ, MN; Data curation: XZ, MN; Formal analysis: XZ, JS, KT; Software: XZ, MN; Validation: XZ, JS, KT; Visualization: XZ; Writing – original draft: XZ, JS, KT; Writing – review \& editing: XZ, JS, KT; Supervision: JS, KT; Project administration: JS, KT; Funding acquisition: JS; Resources: JS. 


\appendix
\section{Stimulus Selection}

From these candidate pairs, we selected 16 target pairs using the Zipf scale of word frequency. The frequency ranges from 1 (lowest frequency) to 7 (highest frequency). Specifically, we conducted an exhaustive search and computed the following measures for each pair:
    \begin{enumerate}
        \item Minimization of the range between the upper and lower Zipf scores for our high-frequency words.
        \item Minimization of the range between the upper and lower Zipf scores for our low-frequency words.
        \item Maximization of the mean frequency difference across all items (the mean frequency of all high-frequency words minus the mean frequency of all low-frequency words).
        \item Maximization of the minimum frequency difference (the minimum frequency of all high-frequency words minus the maximum frequency of all low-frequency words).
    \end{enumerate}




\bibliography{tang}




\end{document}